%% file: main.tex
\documentclass[conference]{IEEEtran}
\IEEEoverridecommandlockouts

\usepackage{cite}
\usepackage{amsmath,amssymb,amsfonts}
\usepackage[dvipdfmx]{graphicx}
\usepackage{textcomp}
\usepackage{xcolor}
\usepackage{multirow}
\input{notations}

\def\BibTeX{{\rm B\kern-.05em{\sc i\kern-.025em b}\kern-.08em
    T\kern-.1667em\lower.7ex\hbox{E}\kern-.125emX}}
\begin{document}

\title{Fast Moving Natural Evolution Strategy\\ for High-Dimensional Problems}

\author{\IEEEauthorblockN{Masahiro Nomura}
\IEEEauthorblockA{\textit{Tokyo Institute of Technology} \\
Tokyo, Japan \\
nomura.m.ad@m.titech.ac.jp}
\and
\IEEEauthorblockN{Isao Ono}
\IEEEauthorblockA{\textit{Tokyo Institute of Technology} \\
Tokyo, Japan \\
isao@c.titech.ac.jp}
}

\maketitle

\IEEEpubidadjcol

\begin{abstract}
In this work, we propose a new variant of natural evolution strategies (NES) for high-dimensional black-box optimization problems. The proposed method, CR-FM-NES, extends a recently proposed state-of-the-art NES, Fast Moving Natural Evolution Strategy (FM-NES), in order to be applicable in high-dimensional problems. CR-FM-NES builds on an idea using a restricted representation of a covariance matrix instead of using a full covariance matrix, while inheriting an efficiency of FM-NES. The restricted representation of the covariance matrix enables CR-FM-NES to update parameters of a multivariate normal distribution in linear time and space complexity, which can be applied to high-dimensional problems. Our experimental results reveal that CR-FM-NES does not lose the efficiency of FM-NES, and on the contrary, CR-FM-NES has achieved significant speedup compared to FM-NES on some benchmark problems. Furthermore, our numerical experiments using 200, 600, and 1000-dimensional benchmark problems demonstrate that CR-FM-NES is effective over scalable baseline methods, VD-CMA and Sep-CMA.
\end{abstract}


\section{Introduction}

\IEEEPARstart{T}{his} work focuses on black-box optimization (BBO) problems.
In BBO, we cannot obtain representations of objective functions explicitly;
optimization should be performed with only function evaluation values for solutions, not gradient or other information.
Because evaluating a solution often demands high computational resources, reducing the number of evaluations is critical in BBO.
For example, hyperparameter optimization~\cite{feurer2019hyperparameter}, which is an important task for achieving high performance of machine learning algorithms and can be formlated as BBO, needs a lot of evaluation time for constructing one training model.
Towards solving such real-world BBO problems, many optimization methods including evolutionary algorithms have been actively studied~\cite{bergstra2011algorithms,bergstra2012random,turner2021bayesian,frazier2018tutorial,loshchilov2016cma,nomura2021warm}.

Our particular interest lies on Natural Evolution Strategies (NES) \cite{glasmachers2010exponential,wierstra2014natural,mcgrath2016unscented,schaul2011high,sun2009efficient,yi2009stochastic,cuccu2012block,bensadon2015black,li2020evolution,glasmachers2010natural,schaul2012natural,nomura2022towards,beyer2014convergence}, which shows promising performance in BBO.
In contrast to typical evolutionary algorithms which minimizes an objective function $f({\bf x})$ by seeking the optimal solution ${\bf x}^{\ast}$ directly, NES attempts to find the parameter $\theta^{*}$ of a probability distribution $p({\bf x}|\theta)$ that minimizes the expected evaluation value $J(\theta) = \int f({\bf x})p({\bf x}|\theta) d{\bf x}$ of the solution ${\bf x}$ sampled from the probability distribution $p({\bf x}|\theta)$.
This technique is called \emph{stochastic relaxation}, and has appealing properties in a theoretical view~\cite{muller2021non}.
NES generates solutions according to a current probability distribution $p({\bf x}|\theta)$ and evaluates them in each iteration.
A critical component in NES is the natural gradient~\cite{amari1998natural,amari1998why,pascanu2013revisiting,martens2020new}, which is defined by the steepest descent direction with a sufficiently small Kullback-Leibler divergence~\cite{kullback1951information,malago2015information,ollivier2017information}.
The natural gradient is estimated by using the evaluation values of the solutions, and the parameter $\theta$ of the distribution is updated by using the estimated natural gradient.

Among the many variants of NES, FM-NES~\cite{nomura2021natural} shows good performance in BBO problems.
In addition to the update based on the natural gradient, FM-NES introduces several mechanisms to speed up the search even with a small number of samples, which results in significant speed-up over other NES algorithms~\cite{glasmachers2010exponential,nomura2021distance} and CMA-ES~\cite{hansen2003reducing,hansen2001completely,hansen2006cma,hansen2016cma}.

However, it is difficult to employ FM-NES for optimizing high-dimensional problems due to the high complexity of FM-NES;
the time complexity of FM-NES is $\mathcal{O}(d^3)$ where $d$ is the dimension number, and the space complexity of FM-NES is $\mathcal{O}(d^2)$, which makes it difficult to apply FM-NES to high-dimensional problems.
This is mainly because the number of parameters of the covariance matrix to be updated is $d(d+1) / 2 + 1 \in \Theta(d^2)$.

To address the issue, we propose, in this paper, Cost-Reduction Fast Moving Natural Evolution Strategy (CR-FM-NES), which is a new version of FM-NES that can be utilized for optimizing high-dimensional problems.
CR-FM-NES builds on the sophisticated representation of the covariance matrix proposed in \cite{akimoto2014comparison}.
While CR-FM-NES retains the high speed of FM-NES, it suppresses the time and space complexity linearly with respect to the dimension number.

The rest of the paper is organized as follows.
In Section~\ref{sec:fmnes}, we describe the existing method, FM-NES, and its problem when applied to high-dimensional problems.
In Section~\ref{sec:proposed}, we propose CR-FM-NES, which deals with the issue of FM-NES.
In Section~\ref{sec:experiments}, we perform numerical experiments to demonstrate the effectiveness of the proposed method.
Section~\ref{sec:conclusion} concludes this study.

\section{FM-NES and Its Problem}
\label{sec:fmnes}

\subsection{FM-NES}
Fast Moving Natural Evolution Strategy (FM-NES)~\cite{nomura2021natural} is one of the state-of-the-art BBO algorithms.
FM-NES performs optimization by updating the parameters of the multivariate normal distribution based on the estimated natural gradient.
We give an overall procedure of FM-NES as follows.


{\bf Step 1.} Initialize the parameters of the multivariate normal distribution $\mathcal{N}({\bf m}^{(0)}, \sigma^{(0)} {\bf B}^{(0)}(\sigma^{(0)} {\bf B}^{(0)})^{\top})$.
Here, ${\bf m}^{(0)}$ is the mean vector, $\sigma^{(0)}$ is the step size, and ${\bf B}^{(0)}$ is the normalized transformation matrix which holds ${\rm det}({\bf B}) = 1$.
Let the iteration $t = 0$ and the evolution path ${\bf p}_{\sigma}^{(0)} = {\bf 0}$.

{\bf Step 2.} Generate $\lambda$ solutions by the \emph{antithetic variates method}~\cite{fukushima2011proposal}; that is, the solutions are generated according to the following equations: ${\bf x}_{2i-1} = {\bf m}^{(t)} + \sigma^{(t)}{\bf B}^{(t)} {\bf z}_{2i-1}, {\bf x}_{2i} = {\bf m}^{(t)} + \sigma^{(t)} {\bf B}^{(t)} {\bf z}_{2i}, {\bf z}_{2i-1} \sim \mathcal{N} ({\bf 0}, {\bf I}), {\bf z}_{2i} = - {\bf z}_{2i-1} (i=1, \cdots, \lambda / 2)$, where $\lambda$ is a positive even number.

{\bf Step 3.} Sort the generated solutions by their evaluation values.

{\bf Step 4.} Update the evolution path~\cite{hansen1996adapting,hansen2006cma} ${\bf p}_{\sigma}^{(t)}$:
\begin{align}
\label{p_sigma_update}
{\bf p}_{\sigma}^{(t+1)} &= (1 - c_{\sigma}) {\bf p}_{\sigma}^{(t)} + \sqrt{c_{\sigma}(2 - c_{\sigma})\mu_{\mathrm{eff}}} \sum_{i=1}^{\lambda} w_i^{\mathrm{rank}} {\bf z}_{i:\lambda},
\end{align}
where ${\bf z}_{i:\lambda}$ means that ${\bf x}_i$ is the $i$-th best solution among the $\lambda$ ones.
The learning rate $c_{\sigma}$, the weight $w_i^{\rm rank}$, and $\mu_{\rm eff}$ are given by
\begin{align}
c_\sigma &= \frac{\mu_{\mathrm{eff}} + 2}{d + \mu_{\mathrm{eff}} + 5}, \\
w_i^{\mathrm{rank}}  &= \frac{\hat{w}_i^{\mathrm{rank}}} {\sum_{j=1}^\lambda \hat{w}_j^{\mathrm{rank}}} - \frac{1}{\lambda}, \label{eq:weight-rank} \\
\hat{w}_i^{\mathrm{rank}} &= \max \left(0, \ln\left(\frac{\lambda}{2} + 1 \right) - \ln{(i)} \right), \\
\label{eq:mueff}
\mu_{\mathrm{eff}} &= 1 / \sum_{i=1}^{\lambda}\left( w_i^{\mathrm{rank}} + \frac{1}{\lambda}\right)^2.
\end{align}

{\bf Step 5.} Set the search phase to ``movement'' if $\mathbb{E} [\| \mathcal{N} ({\bf 0}, {\bf I}) \|] \leq \|{\bf p}_\sigma^{(t+1)} \|$. 
Set the search phase to ``stagnation'' if $0.1 \mathbb{E} [\| \mathcal{N} ({\bf 0}, {\bf I}) \|] \leq \|{\bf p}_\sigma^{(t+1)} \| < \mathbb{E} [\| \mathcal{N} ({\bf 0}, {\bf I}) \|]$.
Otherwise, set the search phase to ``convergence''.

{\bf Step 6.} Set the weight to $w_i = w_i^{\rm dist}$ if the search phase is ``movement''.
Otherwise, set the weight to $w_i = w_i^{\rm rank}$.
Define the distance weight $w_i^{\rm dist}$ as
\begin{align}
\label{eq:weight-dist}
w_i^{\mathrm{dist}}  &= \frac{\hat{w}_i^{\mathrm{rank}}\hat{w}_i^{\mathrm{dist}}}{\sum_{j=1}^\lambda \hat{w}_j^{\mathrm{rank}}\hat{w}_j^{\mathrm{dist}}} - \frac{1}{\lambda}, \\
\label{w_dist_hat}
\hat{w}_i^{\mathrm{dist}} &= \exp\left(\alpha \| {\bf z}_i \| \right),
\end{align}
\noindent
where $\alpha$ is the distance weight parameter. See \cite{nomura2021distance} for how to calculate $\alpha$.

{\bf Step 7.} Set the learning rates $\eta_{\sigma} = \eta_{\sigma}^{\mathrm{move}}$ and $\eta_{\bf B} = \eta_{\bf B}^{\mathrm{move}}$ if the search phase is ``movement''.
If it is ``stagnation'', $\eta_{\sigma} = \eta_{\sigma}^{\mathrm{stag}}$ and $\eta_{\bf B} = \eta_{\bf B}^{\mathrm{stag}}$.
If it is ``convergence'', $\eta_{\sigma} = \eta_{\sigma}^{\mathrm{conv}}$ and $\eta_{\bf B} = \eta_{\bf B}^{\mathrm{conv}}$.
Set $\eta_{\bf m} = 1.0$.
For these learning rates, the recommended values are presented in~\cite{nomura2021distance}.

{\bf Step 8.} Estimate the natural gradients:
\begin{align}
  \begin{split}
  {\bf G}_{\bf M} &= \sum_{i=1}^{\lambda} w_{i} ({\bf z}_{i:\lambda} {\bf z}_{i:\lambda}^{\top} - {\bf I}), G_{\sigma} = \mathrm{Tr} ({\bf G}_{\bf M}) / d, \\
  {\bf G}_{\bf B} &= {\bf G}_{\bf M} - G_{\sigma} {\bf I}, {\bf G}_{\delta} = \sum_{i=1}^{\lambda} w_{i}{\bf z}_{i:\lambda},
  \end{split}
\end{align}
where ${\bf I} \in \mathbb{R}^{d \times d}$ is the $(d\times d)$ identity matrix.

{\bf Step 9.} Update the parameters of the distribution:
\begin{align}
  \label{sigma_update}
  \sigma^{(t+1)} &= \sigma^{(t)} \exp (\eta_{\sigma} G_{\sigma}/2), \\
  \label{m_update}
  {\bf m}^{(t+1)} &= {\bf m}^{(t)} + \eta_{\bf m} \sigma^{(t)} {\bf B}^{(t)} {\bf G}_{\delta}, \\
  \label{B_update}
  {\bf B}^{(t+1)} &= {\bf B}^{(t)} \exp(\eta_{\bf B} {\bf G}_{\bf B}/2).
\end{align}
In addition, update the evolution path ${\bf p}_c^{(t)}$:
\begin{align}
    {\bf p}_{c}^{(t+1)} &= (1-c_{c}) {\bf p}_{c}^{(t)} + \sqrt{c_{c}(2-c_{c})\mu_{\mathrm{eff}}} {\bf B}^{(t)}{\bf G}_{\delta}.
\end{align}

{\bf Step 10.} Emphasize the expansion of the distribution when the search phase is ``movement'':
\begin{align}
  \label{eq:expand}
  {\bf Q} &= (\gamma-1)\sum_{i=1}^d {\mathbb I}(\tau_i > 0) {\bf e}_i {\bf e}_i^{\top} + {\bf I}, \\
  \label{expand-B}
  {\bf B}^{(t+1)} &\gets \mathbf{Q} {\bf B}^{(t+1)}/ \sqrt[d]{{\rm det}({\bf Q})}, \\
  \label{expand-sigma}
  \sigma^{(t+1)} &\gets \sigma^{(t+1)}\sqrt[d]{{\rm det}({\bf Q})},
\end{align}
where $\mathbf{Q} \in \mathbb{R}^{d \times d}$ is a matrix for expanding the normalized transformation one.
$\gamma$ is an expansion rate.
And, $\tau_i (i=1, \ldots, d)$ is a change of a second moment in each direction from ${\bf B}^{(t)} {\bf B}^{(t)^{\top}}$ to ${\bf B}^{(t+1)} {\bf B}^{(t+1)^{\top}}$. The calculation of $\tau_i (i \in \{1, \cdots, d \})$ is defined as
\begin{align}
  \tau_i = \frac{ {\bf e}_i^{\top}({\bf B}^{(t+1)} {\bf B}^{(t+1)^{\top}} - {\bf B}^{(t)} {\bf B}^{(t)^{\top}}) {\bf e}_i}{ {\bf e}_i^{\top} {\bf B}^{(t)} {\bf B}^{(t)^{\top}} {\bf e}_i},
\end{align}
where $\{ {\bf e}_i \}_{i=1}^d$ are the eigenvectors of the normalized covariance matrix, ${\bf B}^{(t)} {\bf B}^{(t)^{\top}}$.
If $\tau_i>0$, the distribution is expanded in a direction of ${\bf e}_i$ due to the above equations.
The update equation of the expansion ratio $\gamma$ is the following.
\begin{align}
  \gamma &\gets \max \Bigl( (1 - c_\gamma)\gamma + c_\gamma \sqrt{1+d_\gamma\tau}, 1\Bigr),\\
  \tau &= \max_i \tau_i,
\end{align}
where the recommended values of $d_\gamma \in \mathbb{R}$ and $c_\gamma \in [0, 1]$ are presented in \cite{nomura2021distance}.

{\bf Step 11.} Perform the rank-one update:
\begin{align}
    {\bf B}^{(t+1)} &= {\bf B}^{(t+1)} \exp(c_1 {\bf R}_{\bf B} / 2), \\
    {\bf R}_{\bf B} &= {\bf R} - {\rm Tr}({\bf R}) {\bf I} / d, \\
    {\bf R} &= ({{\bf B}^{(t)}}^{-1} {\bf p}_{c}^{(t+1)}) ({{\bf B}^{(t)}}^{-1} {\bf p}_{c}^{(t+1)})^{\top} - {\bf I},
\end{align}
where $c_1$ is a learning rate.
For $c_1$, the recommended value is presented in \cite{hansen2016cma}.

{\bf Step 12.} Update the iteration $t \gets t+1$ and move to Step 2 when the stopping criterion is not satisfied.

\subsection{Problem of FM-NES}
A drawback of FM-NES is that it is difficult to apply it to high-dimensional problems.
This is because that the time complexity of FM-NES is $\mathcal{O}(d^3)$, and the space complexity of FM-NES is $\mathcal{O}(d^2)$, which makes it difficult to apply FM-NES to high-dimensional problems.

\section{CR-FM-NES}
\label{sec:proposed}

To address the problem of FM-NES, in this section, we propose Cost-Reduction Fast Moving Natural Evolution Strategy (CR-FM-NES), which reduces the time and space complexity of FM-NES.
The time and space complexity of CR-FM-NES is $\mathcal{O} (d)$, which enables it to be applied to high-dimensional problems.
We first describe basic ideas of CR-FM-NES in Section~\ref{sec:basic_notions}, and then describe the details in Section~\ref{sec:update_cr_params}.
In Section~\ref{sec:cr_lr}, we design the learning rate for CR-FM-NES.
We then present an overall procedure of CR-FM-NES in Section~\ref{sec:cr_alg}

\subsection{Basic Ideas}
\label{sec:basic_notions}
To reduce the time and space complexity of FM-NES, in CR-FM-NES, we utilize the representation of the covariance matrix used in VD-CMA~\cite{akimoto2014comparison}.
We define the covariance matrix of the multivariate normal distribution in CR-FM-NES as follows:
\begin{align}
    \label{eq:repr_vd}
    \C = \sigma^2 \D (\I + \v \v^{\top}) \D,
\end{align}
where $\D \in \mathbb{R}^{d \times d}$ is a diagonal matrix, $\v \in \mathbb{R}^d$ is a $d$-dimensional vector, and $\sigma \in \mathbb{R}_{+}$ is a step size.
The important thing is that, by using this representation instead of the full covariance matrix, the number of parameters is reduced from $d(d+1)/2 + 1 \in \Theta (d^2)$ to $2d + 1 \in \Theta (d)$.

There are several methods restricting the representation of the covariance matrix, for example, sep-CMA-ES~\cite{ros2008simple} and R1-NES~\cite{sun2013linear}.
However, the performance of sep-CMA-ES deteriorates on problems with variable dependencies, and the performance of R1-NES deteriorates on ill-conditioned problems.
We can alleviate these issues by employing the representation in Eq. (\ref{eq:repr_vd}).
Note that, however, there exists objective functions that the covariance matrix with Eq. (\ref{eq:repr_vd}) cannot represent, i.e., the covariance matrix cannot approximate the inverse Hessians of objective functions.

The following operations cannot be \emph{naively} transported to CR-FM-NES with keeping the time and space complexity $\O (d)$.
\begin{itemize}
    \item Emphasizing the expansion of the probability distribution
    \item Updating the parameters $\v$ and $\D$
\end{itemize}

To achieve the linear complexity, we do not include emphasizing the expansion of the probability distribution in CR-FM-NES.
We describe how to update the parameter $\v$ and $\D$ with linear complexity in Section \ref{sec:update_cr_params}.

Due to the restrictive representation of the covariance matrix, it is expected that the learning rate of the covariance matrix in CR-FM-NES can be set to higher values than that of FM-NES because the number of the parameters in the covariance matrix is reduced from $d(d+1)/2 + 1 \in \Theta (d^2)$ to $2d + 1 \in \Theta (d)$.
We describe the specific design of the learning rate in Section \ref{sec:cr_lr}.

In the following, we use the notations without redefinition if they have been already introduced in Section~\ref{sec:fmnes} (for example, the weight function $w(\cdot)$ and the learning rates).

\subsection{Parameter Update}
\label{sec:update_cr_params}

\subsubsection{$\v$ and $\D$ Update}
The parameters of the covariance matrix in CR-FM-NES, $\v$ and $\D$, are updated by the estimated natural gradient as follows~\cite{akimoto2014comparison}:
\begin{align}
    \v^{(t+1)} &= \v^{(t)} + \eta_{\B} \sum_{i=1}^{\lambda}w_i\tilde{\nabla}_v\ln{p_{\theta}(\x_{i})} \notag
    \\ & \hspace{10mm}
  + c_1 \tilde{\nabla}_v\ln{p_{\theta}(\m^{(t)} + \sigma^{(t)} \p_c^{(t+1)})}\label{eq:update_v},\\
  {\bf D}^{(t+1)} &= {\bf D}^{(t)} + \eta_{\B} \sum_{i=1}^{\lambda}w_i\tilde{\nabla}_D\ln{p_{\theta}(\x_{i})} \notag
  \\ & \hspace{10mm}
  + c_1 \tilde{\nabla}_D\ln{p_{\theta}(\m^{(t)} + \sigma^{(t)} \p_c^{(t+1)})}    \label{eq:update_D},
\end{align}
where $\tilde{\nabla}_{\t}\ln{p_{\theta}(\cdot)} = \F^{-1}(\t) \nabla_{\t} \ln{p_{\theta}(\cdot)}$.
$\F (\t)$ is the Fisher information matrix, $p_{\theta}(\cdot)$ is the multivariate normal distribution, and $\theta = (\m^{\top} \in \mathbb{R}^d, \sigma \in \mathbb{R}_{+}, \v^{\top} \in \mathbb{R}^d, \theta_{\D}^{\top} \in \mathbb{R}^d)^{\top}$, where $\theta_{\D}$ is a real-valued vector whose $i$th element is the $i$th diagonal element of $\D$.
It should be noted that, however, we do not have to calculate the inverse of the Fisher information matrix explicitly, which will be shown in the next paragraph.

Following~\cite{akimoto2014comparison}, the natural gradients
$\tilde{\nabla}_v\ln{p_{\theta}(\x)}$ and
$\tilde{\nabla}_D\ln{p_{\theta}(\x)}$ are updated as follows:
\begin{align}
\tilde{\nabla}_v\ln{p_{\theta}(\x)} &= \|\v\|^{-1} \tv, \\
\tilde{\nabla}_D\ln{p_{\theta}(\x)} &= \D \s,
\end{align}
where $\s$ and $\tv$ are calculated by the following steps:
\begin{align*}
  {\rm 1.}&\quad {\rm Let}\ \y := {\D^{(t)}}^{-1}(\x - \m^{(t)})/\sigma, \bar{\v} := \v^{(t)} / \|\v^{(t)}\|. \\
  &\quad {\rm Set}\ \s = \y \odot \y - \|\v^{(t)}\|^2 \langle \y,\bar{\v} \rangle \gamma_v^{-1}\y \odot \bar{\v} - \1. \\
  {\rm 2.}&\quad {\rm Let}\ \gamma_v := 1 + \|\v^{(t)}\|^2. \\
  &\quad {\rm Set}\ \tv = \langle \y,\bar{\v} \rangle \y - 2^{-1}(\langle \y,\bar{\v} \rangle^2 + \gamma_v) \bar{\v}. \\
  {\rm 3.}&\quad {\rm Let}\\
  &\quad \alpha_{\rm vd} := \min \Bigl(1, \frac{[\|\v^{(t)}\|^4 + (2-\gamma_{\rm vd})\gamma_v / \max_i (\bar{\bar{\v}}_i)]^{1/2}}{2 + \|\v^{(t)}\|^2} \Bigr),  \\
  &\quad {\rm and}\ \gamma_{\rm vd} := \gamma_v^{-1/2},\ {\rm and}\ \bar{\bar{\v}} := \bar{\v} \odot \bar{\v}.\quad {\rm Update} \\
  &\quad \s \gets \s - \alpha_{\rm vd} \gamma_v^{-1}((2+\|\v^{(t)}\|^2)\bar{\v}\odot \tv - \|\v^{(t)}\|^2 \langle \bar{\v},\tv \rangle \bar{\bar{\v}}).  \\
  {\rm 4.}&\quad {\rm Let}\ \bar{\V} = \V / \|\v^{(t)}\|, b := -(1-\alpha_{\rm vd}^2)\|\v^{(t)}\|^4 \gamma_v^{-1} + 2\alpha_{\rm vd}^2 \\
  &, {\rm and}\ \H := 2\I - (b + 2\alpha_{\rm vd}^2)\bar{\V}^2.\ {\rm Update}\ \\
  &\quad \s \gets \H^{-1}\s - (1 + b\langle \bar{\bar{\v}}, \H^{-1}\bar{\bar{\v}}\rangle )^{-1} b\langle \s, \H^{-1}\bar{\bar{\v}}\rangle \H^{-1}\bar{\bar{\v}}. \\
  {\rm 5.}&\quad {\rm Update}\ \tv \gets \tv - \alpha_{\rm vd} [(2+\|\v^{(t)}\|^2)\bar{\v}\odot \s - \langle \s,\bar{\bar{\v}} \rangle \bar{\v}  ]. 
\end{align*}
Here, $\V$ is a diagonal matrix whose diagonal elements are composed of $\v$.
$\odot$ denotes an operation which performs element-wise product.
We emphasize that all the above operations can be performed in linear complexity.
Note that we employ a corrected version of $\H$ which is different from that in \cite{akimoto2014comparison};
we leave the detailed derivation in Appendix.

After updating $\v$ in Eq.~\ref{eq:update_v} and $\D$ in Eq.~\ref{eq:update_D}, we normalize $\D$ to keep the determinant of $\D (\I + \v \v^{\top}) \D$.
This implies that the determinant of the covariance matrix is affected only by the $\sigma$ update, which will be described in the next section.

\subsubsection{$\sigma$ Update}
Similarly to FM-NES, we update the step-size $\sigma$ by using the estimated natural gradient as follows.
Note that this operation also can be performed in linear time.
\begin{align}
  \sigma^{(t+1)} &= \sigma^{(t)} \cdot \exp( \eta_{\sigma}/2 \cdot G_{\sigma}),\\
  G_{\sigma} &= {\rm Tr}\Bigl( \sum_{i=1}^{\lambda}w_i(\z_{i}\z_{i}^{\top} - {\bf I}) \Bigr) / d.
\end{align}

\subsection{Learning Rates Design}
\label{sec:cr_lr}
\begin{figure}[tb]
  \centering
  \includegraphics[width=70mm]{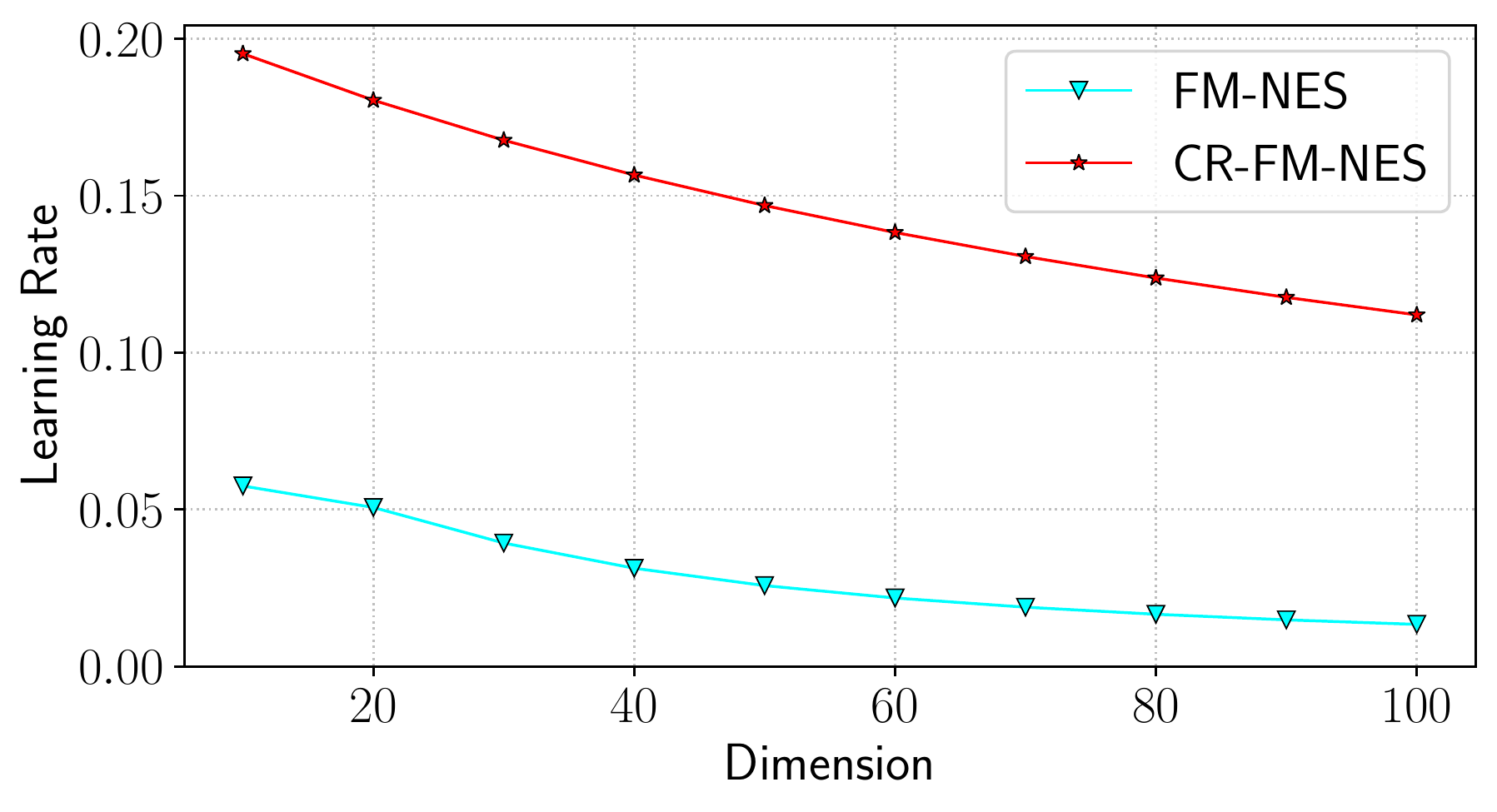}
  \caption{Comparing the learning rate for the covariance matrix used in FM-NES (cyan) and CR-FM-NES (red). The horizontal axis represents the dimension number. The vertical axis represents the learning rate of the covariance matrix. For the learning rate of FM-NES, the value set when the search phase is the \emph{movement} phase~\cite{nomura2021natural} is used.}
  \label{fig:lr_fm_and_cr}
\end{figure}

We design, in this section, learning rates with restrictring the representation of the covariance matrix.
This definition of the learning rate $c_1$ used to the rank-one update is the same as that used in \cite{akimoto2014comparison}, i.e., 
\begin{align}
    c_1 &= \frac{d-5}{6} c_1^{\rm cma},\\
    c_1^{\rm cma} &= \frac{2}{(d+1.3)^2 + \mu_{\rm eff}}.
\end{align}
Additionally, we set the learning rate $\eta_{\B}$ used for $\v$ and $\D$ to
\begin{align}
    \eta_{\B} = {\rm tanh} \left( \frac{\min(0.02 \lambda, 3 \ln(d)) + 5}{0.23 d + 25} \right).
\end{align}
We determined this equation by fitting a parametric model given the fixed $c_1$ defined above.
Figure~\ref{fig:lr_fm_and_cr} shows a comparison of the learning rate for the covariance matrix used in FM-NES and CR-FM-NES when we vary the dimension number $d \in \{ 10, 20, \cdots, 100 \}$ and set $\lambda = 20$.
We can see that the learning rate used in CR-FM-NES is much larger than the one used in FM-NES.
For other parameters in CR-FM-NES, the recommended values of FM-NES~\cite{nomura2021natural} are used .

\subsection{Algorithm of CR-FM-NES}
\label{sec:cr_alg}
Algorithm~\ref{alg:provd} shows the overall procedure of CR-FM-NES.
Note that ${\bf E}$ is the set of positive even integers.
In lines 3-7, candidate solutions are generated from the multivariate normal distribution.
In line 8, the candidate solutions are sorted with respect to their evaluation values.
In line 9, updating the evolution path $\p_{\sigma}$ is performed.
In line 10-11, switching the weight function and the learning rates is performed by using the norm of the evolution path.
In line 12, updating the evolution path $\p_{c}$ is performed.
In lines 13-19, updating the parameters of the multivariate normal distribution is performed based on the estimated natural gradient and the evolution path.
Note that $\D$ is normalized to keep the determinant of $\D (\I + \v \v^{\top}) \D$ in lines 16-17.

\begin{algorithm}
\caption{CR-FM-NES}
\label{alg:provd}
\begin{algorithmic}[1]
\Require $d \in \mathbb{N},\ f: \mathbb{R}^d \to \mathbb{R}, \lambda \in {\bf E}, \m^{(0)} \in \mathbb{R}^d, {\bf D}^{(0)} \in \mathbb{R}^{d\times d}, \v^{(0)} \in \mathbb{R}^d, \sigma^{(0)} \in \mathbb{R}_{+} $
\State $t = 0$, $\p_{\sigma}^{(0)} = \0$, $\p_c^{(0)} = \0$, $\Upsilon = \mathbb{E}\|\mathcal{N}(\0,\mbox{\bf I})\|$
\While{stopping criterion not met}
  \For{$i \in \{ 1, \cdots, \lambda / 2 \}$}
    \State $\z_{2i-1} \sim \mathcal{N}({\bf 0}, \I), \z_{2j} = -\z_{2j-1}$
    \State $\y_{2i-1} = \z_{2i-1}+(\sqrt{1+\|\v^{(t)}\|^2}-1)\langle \z_{2i-1},\bar{\v}\rangle \bar{\v}, \y_{2i} = \z_{2i}+(\sqrt{1+\|\v^{(t)}\|^2}-1)\langle \z_{2i},\bar{\v}\rangle \bar{\v}$
    \State $\x_{2i-1} = \m^{(t)} + \sigma^{(t)} {\bf D}^{(t)}\y_{2i-1}, \x_{2i} = \m^{(t)} + \sigma^{(t)} {\bf D}^{(t)}\y_{2i} $
  \EndFor
  \State sort $\{ (\z_i, \y_i, \x_i)\}$ with respect to $f(\x_i)$
  \State $\p_{\sigma}^{(t+1)} = (1-c_{\sigma})\p_{\sigma}^{(t)} + \sqrt{c_{\sigma}(2-c_{\sigma})\mu_{\mathrm{eff}}}\sum_{i=1}^{\lambda}w_i^{\rm rank}\z_{i}$
  \State {\bf if} $ \|\p_{\sigma}^{(t+1)}\| \geq \Upsilon$ {\bf then} $w_i = w_i^{\rm dist}$ {\bf else} $w_i = w_i^{\rm rank}$
  \State {\bf if} $ \|\p_{\sigma}^{(t+1)}\| \geq \Upsilon$ {\bf then} $\eta_{\sigma} = \eta_{\sigma}^{\rm move}$ {\bf else if} $ \|\p_{\sigma}^{(t+1)}\| \geq 0.1 \Upsilon$ {\bf then}  $\eta_{\sigma} = \eta_{\sigma}^{\rm stag}$ {\bf else} $\eta_{\sigma} = \eta_{\sigma}^{\rm conv}$

  \State $\p_{c}^{(t+1)} = (1-c_{c})\p_{c}^{(t)} + \sqrt{c_{c}(2-c_{c})\mu_{\rm eff}}\sum_{i=1}^{\lambda}w_i(\x_{i}-\m^{(t)})/\sigma^{(t)}$
  \State $\m^{(t+1)} = \m^{(t)} + \eta_{m}\sum_{i=1}^{\lambda}w_i (\x_{i} - \m^{(t)}) $

  \State $\v^{(t+1)} = \v^{(t)} + \eta_\B \sum_{i=1}^{\lambda}w_i\tilde{\nabla}_v\ln{p_{\theta}(\x_{i})} + c_1 \tilde{\nabla}_v\ln{p_{\theta}(\m^{(t)} + \sigma^{(t)} \p_c^{(t+1)})}$
  \State ${\bf D}^{(t+1)} = {\bf D}^{(t)} + \eta_\B \sum_{i=1}^{\lambda}w_i\tilde{\nabla}_D\ln{p_{\theta}(\x_{i})} + c_1 \tilde{\nabla}_D\ln{p_{\theta}(\m^{(t)} + \sigma^{(t)} \p_c^{(t+1)})}$

  \State ${\rm detA}^{(t+1)} = {\small {\rm det}({\bf D}^{(t+1)}({\bf I}+\v^{(t+1)}{\v^{(t+1)}}^{\top}){\bf D}^{(t+1)}) }$
  \State ${\bf D}^{(t+1)} \leftarrow {\bf D}^{(t+1)} / \sqrt[2d]{{\rm detA}^{(t+1)}}$
  
  \State $G_{\sigma} = {\rm Tr}\Bigl( \sum_{i=1}^{\lambda}w_i(\z_{i}\z_{i}^{\top} - {\bf I}) \Bigr) / d$
  \State $\sigma^{(t+1)} = \sigma^{(t)} \cdot \exp( \eta_{\sigma}/2 \cdot G_{\sigma})$
  \State $t \leftarrow t + 1$
\EndWhile

\end{algorithmic}
\end{algorithm}

\section{Experiments}
\label{sec:experiments}
In this work, we experiments with benchmark problems in order to investigate the following research questions (RQs).

\begin{itemize}
    \item [\textbf{RQ1.}] How is the performance of CR-FM-NES, compared to FM-NES?
    \item [\textbf{RQ2.}] Is CR-FM-NES more efficient than baseline methods for high-dimension problems?
\end{itemize}

We first describe the experimental settings in Section~\ref{sec:exp_setup}.
In Section~\ref{sec:exp_low}, we compare FM-NES with CR-FM-NES on $80$-dimensional problems (\textbf{RQ1}).
We then compare CR-FM-NES with other baseline methods (\textbf{RQ2}) in Section~\ref{sec:exp_high}.
The code for running the proposed method is available at \href{https://github.com/nomuramasahir0/crfmnes}{\textbf{https://github.com/nomuramasahir0/crfmnes}}.

\subsection{Settings}
\label{sec:exp_setup}
The definitions of benchmark problems which we used in the experiment are shown in Table~\ref{tab:benchmark}.
For the $k$-Tablet function, we set $k = d / 4$.
For the Sphere, $k$-Tablet, Ellipsoid, and Rastrigin functions, we set the initial distribution to $\m^{(0)} = (3, \cdots, 3)^{\top}, \sigma^{(0)} = 2.0, \B^{(0)} = {\bf I}$.
For the Rosenbrock function, we set $\m^{(0)} = (0, \cdots, 0)^{\top}, \sigma^{(0)} = 0.5, \B^{(0)} = {\bf I}$.

As the performance metrics, we employ the average number of evaluations until the best evaluation value $f(\x_{\rm best})$ reaches the target objective value $10^{-10}$ over successful trials divided by the success rate~\cite{auger2005restart}.
A trial is judged to be successful if the target objective value is obtained within the maximum number of evaluations, $5d \times 10^4$.
Based on preliminary experiments, we determined the settings of the population size $\lambda$ as follows: 
For all the Sphere, $k$-Tablet, Ellipsoid, and Rosenbrock functions, we employ $\lambda = \lambda_{\rm def}, 2 \lambda_{\rm def}, 3 \lambda_{\rm def}, 4 \lambda_{\rm def}, 5 \lambda_{\rm def}$, where
\begin{equation}
  \lambda^{\rm def} = \begin{cases}
    4 + \lfloor 3\ln{(d)} \rfloor & (\lfloor 3\ln{(d)} \rfloor \ {\rm mod}\ 2 = 0), \\
    5 + \lfloor 3\ln{(d)} \rfloor & ({\rm otherwise}).
    \end{cases}
\end{equation}
Note that $\lambda$ is set to be an even number to use the antithetic sampling method.
In this setting, we obtain $\lambda^{\rm def} = 18, 20, 24$, and $24$ for the dimension number $d = 80, 200, 600$, and $1000$, respectively.
In addition, Table~\ref{tab:rastrigin_lambda} shows the population sizes used in the experiments on the Rastrigin function.
A trial number (that is, the number of times that the experiment is repeated to calculate the average number of evaluations and the success rate) is set to $30$ for the Rastrigin function, and $10$ for the other functions.

\begin{table}[t]
  \centering
  \caption{Definition of benchmark problems. We set $k=d / 4$ for the $k$-Tablet function in our experiments.}
  \label{tab:benchmark}
  \begin{tabular}{ll}
    \bottomrule
    Name & Definition \\
    \hline \hline
    Sphere & $f_{{\rm sph}}(\x) = \sum_{i=1}^{d} \x_i^2$  \\
    $k$-Tablet & $f_{k{\rm tab}}(\x) = \sum_{i=1}^{k} \x_i^2 + \sum_{i=k+1}^d (100 \x_i)^2$  \\
    Ellipsoid & $f_{\rm ell}(\x) = \sum_{i=1}^{d} (1000^{\frac{i-1}{d-1}} \x_i)^2$ \\
    Rosenbrock & $f_{\rm rosen}(\x) = \sum_{i=1}^{d-1} 100 (\x_{i+1} - \x_i^2)^2 + (\x_i - 1)^2$ \\
    Rastrigin & $f_{\rm rast}(\x) = 10d + \sum_{i=1}^d (\x_i^2 - 10 \cos(2\pi \x_i))$ \\
    \bottomrule
  \end{tabular}
\end{table}

\begin{table}[t]
  \centering
  \caption{\small Configurations of the population size $\lambda$ used for the experiments on the Rastrigin function.}\label{rastrigin_lambda}
  \label{tab:rastrigin_lambda}
  \begin{tabular}{|l|l|}
    \hline
     dimension $d$ & population size $\lambda$ \\
     \hline \hline
80 & $20d, 22d, 24d, 26d, 28d$
\\ \hline
 200 & $12d, 14d, 16d, 18d, 20d$
\\ \hline
 600 & $6d, 7d, 8d, 9d, 10d$
\\ \hline
 1000 & $4d, 4.5d, 5d, 5.5d, 6d$
\\ \hline
  \end{tabular}
\end{table}

\subsection{Comparison to FM-NES}
\label{sec:exp_low}
\begin{figure*}[tb]
  \centering
  \includegraphics[width=180mm]{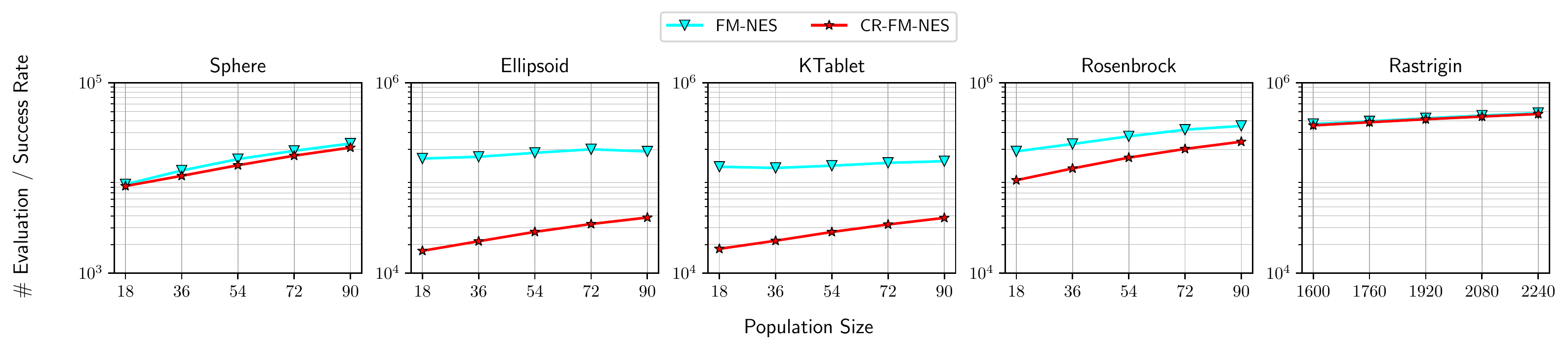}
  \caption{Performance comparison of FM-NES (cyan) with CR-FM-NES (red) on $80$-dimensional benchmark problems. The x-axis shows the population size and the y-axis shows the average number of evaluations divided by the success rate. Note that the smaller the value is, the better it is.}
  \label{fig:fm_vs_cr}
\end{figure*}

To compare the performance of FM-NES and that of CR-FM-NES, we conduct experiments on $80$-dimensional benchmark problems.
Figure~\ref{fig:fm_vs_cr} shows the results in each benchmark problem.
For the Ellipsoid, the $k$-Tablet, and the Rosenbrock functions, CR-FM-NES has achieved a significant speedup compared to FM-NES.
We argue that this is because the number of parameters of the covariance matrix in CR-FM-NES is $2d + 1 \in \Theta (d)$, which enables us to set the learning rates much higher than those in FM-NES, as described in Section~\ref{sec:cr_lr}.
In addition, the result in the Rastrigin function implies that CR-FM-NES does not lose its efficiency even in multimodal problems.
Note that, however, CR-FM-NES will fail to perform optimization in problems whose inverse Hessian of the objective function cannot be approximated by the covariance matrix represented in Eq. (\ref{eq:repr_vd}), as illustrated in \cite{akimoto2014comparison}.

\begin{figure}[tb]
  \centering
  \includegraphics[width=70mm]{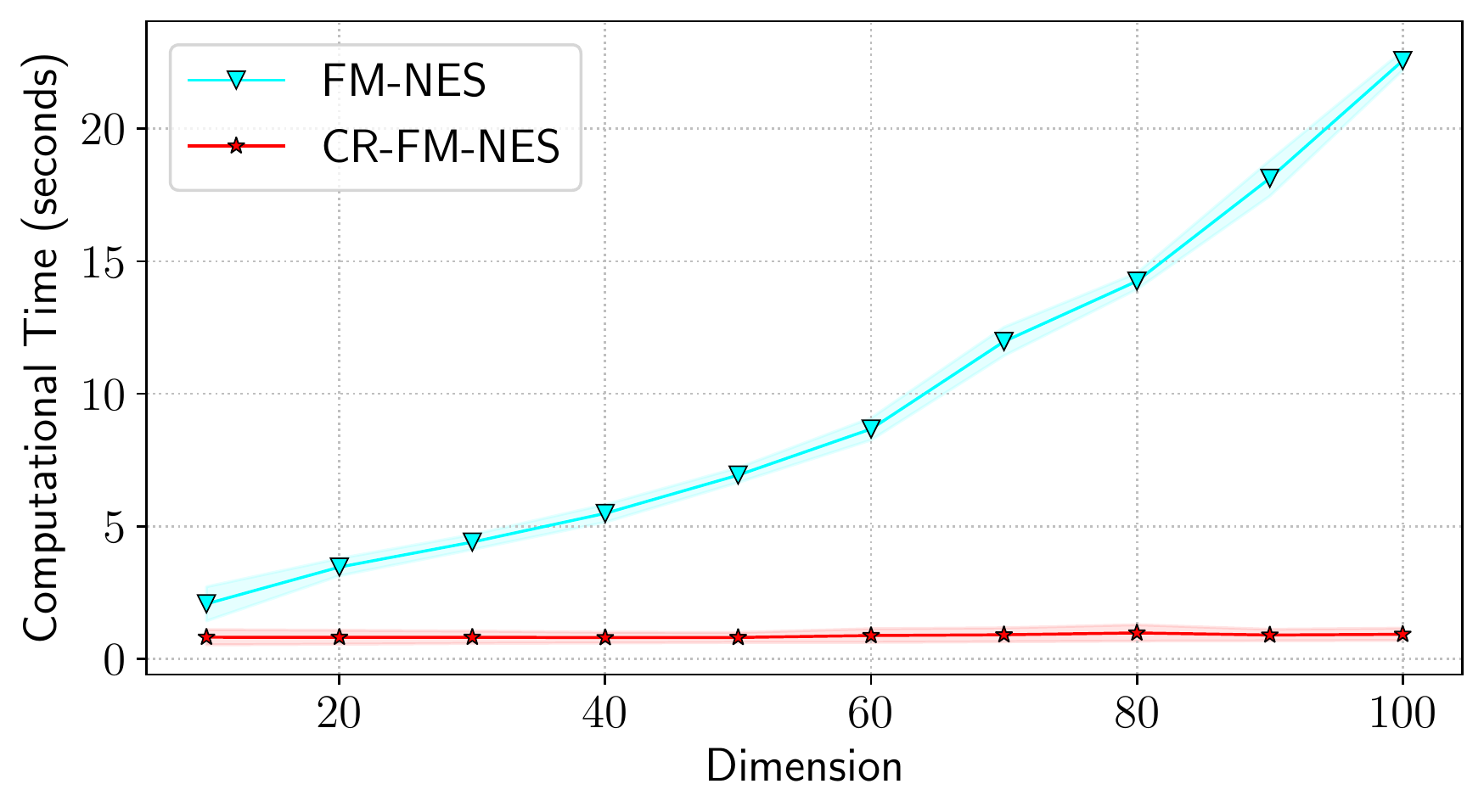}
  \caption{Comparison of the computational time between FM-NES (cyan) and CR-FM-NES (red). The x-axis shows the dimension number and the y-axis shows the average (line) and standard deviation (shadow) of time required to execute 1000 iterations for each method over 30 trials. Note that the standard deviations are small and then hardly visible.}
  \label{fig:time_fm_and_cr}
\end{figure}

We additionally investigate the computational time of both methods.
The CPU is Intel Xeon E5-2680 V4 Processor (2.4GHz) and the memory is 7.5GB.
The OS is SUSE Linux Enterprise Server 12 SP4.
All the code used in the experiment is implemented by Python and its version when executed is 3.6.3.
Figure~\ref{fig:time_fm_and_cr} shows the computational time required to execute 1000 iterations for each method.
We vary the dimension number $d \in \{ 10, 20, \cdots, 100 \}$ and set $\lambda = 20$.
We can confirm that the computational time of FM-NES increases rapidly as the dimension number increases, while that of CR-FM-NES is fairly small.

\begin{figure*}[tb]
  \centering
  \includegraphics[width=180mm]{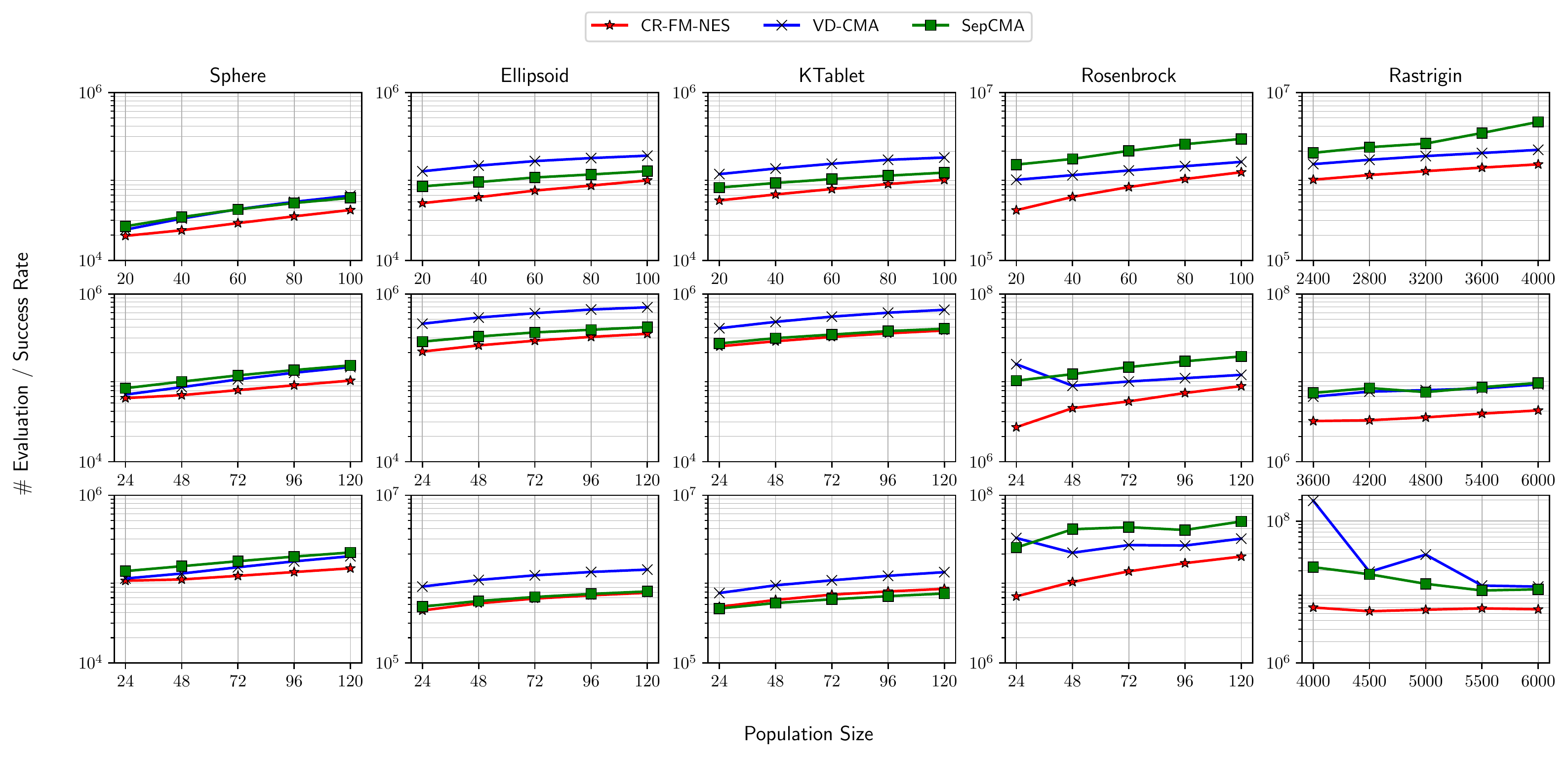}
  \caption{Performance comparison of CR-FM-NES (red), VD-CMA (blue), and Sep-CMA (green) on high-dimensional problems. (Top) $d=200$, (Middle) $d=600$, (Bottom) $d=1000$. The x-axis shows the population size and the y-axis shows the average number of evaluations divided by the success rate. Note that the smaller the value is, the better it is.}
  \label{fig:highdim}
\end{figure*}

\subsection{Comparison in high-dimensional problems}
\label{sec:exp_high}
In this section, to verify the effectiveness of CR-FM-NES in high-dimensional problems, we compare CR-FM-NES with VD-CMA~\cite{akimoto2014comparison} and Sep-CMA~\cite{ros2008simple}.
Although VD-CMA uses the same representation of the covariance matrix as CR-FM-NES, its procedure is fairly different as VD-CMA is based on CMA-ES.
Sep-CMA is also based on CMA-ES, but restricts the covariance matrix to a diagonal matrix.
We do not include R1-NES for the baseline methods because Akimoto et al.~\cite{akimoto2014comparison} have been reported that VD-CMA shows clearly better performance than R1-NES.
We experiment on problems with dimension $d=200, 600,$ and $1000$.

Figure~\ref{fig:highdim} shows the results of the experiment.
For the Ellipsoid and the $k$-Tablet functions where the inverse Hessian can be approximated by even a diagonal matrix, the performance of CR-FM-NES is competitive with that of Sep-CMA.
In contrast, for the Rosenbrock function where the inverse Hessian \emph{cannot} be approximated by a diagonal matrix, the efficiency of CR-FM-NES increases substantially compared to Sep-CMA, and it also shows better performance than VD-CMA.
We believe that the performance difference between CR-FM-NES and VD-CMA is originated from the one in FM-NES and CMA-ES;
CR-FM-NES inherits the efficiency of FM-NES, which leads to good performance in high-dimensional problems.
From the results of the Rastrigin function, we might argue that CR-FM-NES is efficient even in high-dimensional multimodal problems.
Further investigations in this direction are left for future work.

\section{Conclusion}
\label{sec:conclusion}
In this work, we introduced a new NES variant for high-dimensional problems, Cost-Reduction Fast Moving Natural Evolution Strategy (CR-FM-NES).
CR-FM-NES extends FM-NES, a recently proposed state-of-the-art NES algorithm, to be applicable in high-dimensional problems.
To do so, CR-FM-NES uses the representation of the covariance matrix proposed in \cite{akimoto2014comparison}, instead of a full covariance matrix.
Our experimental results suggest that compared to FM-NES, CR-FM-NES has achieved significant speedup on some problems or at least achieved competitive performance.
Furthermore, CR-FM-NES showed favorable performance over VD-CMA~\cite{akimoto2014comparison} and Sep-CMA~\cite{ros2008simple}.

The main limitation of the proposed method lies on the restricted representation of the covariance matrix.
As noted in \cite{akimoto2014comparison}, CR-FM-NES will fail optimization when the covariance matrix cannot approximate the inverse Hessian of the objective function.
Therefore, estimating the problem complexity in online and switching the level of the representation of the covariance matrix, such as online model selection~\cite{akimoto2016online}, is an important future direction.

Additionally, application of CR-FM-NES to real-world problems such as machine learning is also an exciting direction.
For example, in optimizing parameters of scalable machine learning algorithms, it is often necessary to solve high-dimensional and multimodal optimization problems~\cite{sun2019survey,sra2012optimization,li2018visualizing}.
From our experimental results that CR-FM-NES is efficient even in high-dimensional and multimodal problems, we believe that there is a room that CR-FM-NES can have an active role in real-world applications including machine learning.
Furthermore, not only comparison with evolutionary algorithms, but also comparison with gradient-based method such as stochastic gradient descent~\cite{bottou2012stochastic,ruder2016overview}, which has already become a de facto standard in machine learning, is important to expand the scope of application of the method.

\section*{Acknowledgement}
The authors thank anonymous reviewers for their helpful comments.
This work was partially supported by JSPS KAKENHI Grant Number JP20K11986.

\bibliographystyle{IEEEtran}
\bibliography{ref}

\appendix

\textbf{Derivation of $\H$ (Modification of Lemma 3.4 in \cite{akimoto2014comparison})}:
\label{app:schur}
In this appendix, we prove that when the Fisher information matrix is defined as ${\rm diag} (\mathcal{I}_{m}, \mathcal{I}_{C}^{(\alphavd)})$, where $\mathcal{I}_{m} = \sigma^{-2} \left(\D (\I + \v \v^{\top}) \D \right)^{-1}$ and $\mathcal{I}_{C}^{(\alphavd)} = 
  \begin{bmatrix}
    \mathcal{I}_{v,v} & \alphavd \mathcal{I}_{v,D} \\
    \alphavd \mathcal{I}_{D,v} & \mathcal{I}_{D,D}
  \end{bmatrix}$,
the Schur complement of $\mathcal{I}_{v,v}$ holds
$S_{v,v} = \D^{-1}(\H + b\bar{\bar{\v}} \bar{\bar{\v}}^{\rm T})\D^{-1}$.
Here, 
$\bar{\v} = \v / \|\v\|$,
$\gamma_v = 1 + \|\v\|^2$, 
$\gamma_{\rm vd} := \gamma_v^{-1/2}$,
$\alpha_{\rm vd} = \min \Bigl(1, \frac{[\|\v\|^4 + (2-\gamma_{\rm vd})\gamma_v / \max_i (\bar{\bar{\v}}_i)]^{1/2}}{2 + \|\v\|^2} \Bigr)$, 
$\bar{\V} = \V / \|\v\|$, 
$b = -(1-\alphavd^2)\|\v\|^4 \gamma_v^{-1} + 2\alphavd^2$,
$\H = 2\I - (b + 2\alphavd^2)\bar{\V}^2$.
As introduced in Section~\ref{sec:proposed}, $\V$ is a diagonal matrix whose diagonal elements are composed of $\v$.
Note that the resulting $\H$ is different from the value derived in \cite{akimoto2014comparison}.
For a fair comparison, in our experiments, we use this definition of $\H$ for not only CR-FM-NES (our proposal) but also VD-CMA.
For the exact value of $\mathcal{I}_{v,v}, \mathcal{I}_{v,D}$, and $\mathcal{I}_{D,D}$, see the Lemma 3.2 in \cite{akimoto2014comparison}.

\begin{align*}
  S_{v,v} &= \mathcal{I}_{D,D} - \alphavd^2 \mathcal{I}_{D,v}\mathcal{I}_{v,v}^{-1} \mathcal{I}_{v,D} \\
    &= \gamma_v^{-1}\D^{-1}[ 2\gamma_v\I + \|\v\|^2\V^2 - \V \v\v^{\rm T}\V ] \D^{-1} \notag \\
      &\hspace{10mm} - \alphavd^2 \frac{1}{1+\|\v\|^2}\D^{-1}\bar{\V} \Bigl[ (2+\|\v\|^2)^2\I \notag \\
      &\hspace{10mm} - (2 + 2\|\v\|^2 + \|\v\|^4) \bar{\v}\bar{\v}^{\rm T} \Bigr] \bar{\V}\D^{-1} \\
    &= \D^{-1} \Bigl[2\I + \frac{\|\v\|^2}{1+\|\v\|^2}\|\v\|^2\bar{\V}^2 \notag \\
    &\hspace{10mm} - \frac{1}{1+\|\v\|^2}\|\v\|^2\bar{\V}\bar{\v}\bar{\v}^{\rm T}\|\v\|^2\bar{\V} \\
      &\hspace{10mm} - \alphavd^2\frac{(2+\|\v\|^2)^2}{1+\|\v\|^2}\bar{\V}^2 \notag \\
      &\hspace{10mm} + \alphavd^2\frac{2+2\|\v\|^2+\|\v\|^4}{1+\|\v\|^2}\bar{\V}\bar{\v}\bar{\v}^{\rm T}\bar{\V} \Bigr]\D^{-1} \\
    &= \D^{-1}\Bigl[2\I + \frac{\|\v\|^4 - \alphavd^2(2+\|\v\|^2)^2}{1+\|\v\|^2}\bar{\V}^2 \notag \\
    &\hspace{3mm} - \frac{\|\v\|^4 - \alpha_{\rm vd}^2(2+2\|\v\|)^2+\|\v\|^4}{1+\|\v\|^2}\bar{\V}\bar{\v}\bar{\v}^{\rm T}\bar{\V} \Bigl]\D^{-1}.
\end{align*}
By letting $b := - (1-\alphavd^2)\|\v\|^4\gamma_v^{-1} + 2\alphavd^2$,
\begin{align*}
  &\frac{\|\v\|^4 - \alphavd^2(2+\|\v\|^2)^2}{1+\|\v\|^2}\bar{\V}^2 \notag \\
    & \hspace{5mm} =  \frac{1}{1+\|\v\|^2}[\|\v\|^4 - \alphavd^2 \cdot 4 \notag \\
    &\hspace{10mm} - \alphavd^2 \cdot 4\|\v\|^2 - \alphavd^2 \|\v\|^4 ] \bar{\V}^2 \\
    & \hspace{5mm} = \Bigl[ \frac{\|\v\|^4}{1+\|\v\|^2}(1-\alphavd^2) \notag \\
    & \hspace{10mm} + \frac{1}{1+\|\v\|^2}(-4\alphavd^2 -4\alphavd^2\|\v\|^2) \Bigr] \bar{\V}^2 \\
    & \hspace{5mm} = \Bigl[ -b + 2\alphavd^2 - 4\alphavd^2 \Bigl( \frac{1+\|\v\|^2}{1+\|\v\|^2} \Bigr) \Bigr] \bar{\V}^2 \\
    & \hspace{5mm} = - (b + 2\alphavd^2) \bar{\V}^2.
\end{align*}
By $\bar{\V}\bar{\v}\bar{\v}^{\rm T}\bar{\V} = \bar{\V}\bar{\v}\bar{\v}^{\rm T}\bar{\V}^{\rm T} = \bar{\bar{\v}}\bar{\bar{\v}}^{\rm T}$,
\begin{align*}
  &\frac{\|\v\|^4 - \alphavd^2(2 + 2\|\v\|^2 + \|\v\|^4)}{1+\|\v\|^2}\bar{\V}\bar{\v}\bar{\v}^{\rm T}\bar{\V} \notag \\
  & \hspace{5mm} = \Bigl[ \frac{\|\v\|^4}{1+\|\v\|^2}(1-\alphavd^2) - 2\alphavd^2 \Bigr] \bar{\bar{\v}}\bar{\bar{\v}}^{\rm T} \\
  & \hspace{5mm} = -b \bar{\bar{\v}}\bar{\bar{\v}}^{\rm T}.
\end{align*}
Therefore, by letting $\H := 2\I - (b+2\alphavd^2)\bar{\V}^2$,
\begin{align*}
  S_{v,v} &= \D^{-1}[ 2\I - (b+2\alphavd^2)\bar{\V}^2 + b\bar{\bar{\v}}\bar{\bar{\v}}^{\rm T} ]\D^{-1} \\
    &= \D^{-1}[\H + b\bar{\bar{\v}}\bar{\bar{\v}}^{\rm T}]\D^{-1}.
\end{align*}

\end{document}

%% file: notations.tex
\usepackage[utf8]{inputenc} 
\usepackage[T1]{fontenc}    
\usepackage{hyperref}       
\usepackage{url}            
\usepackage{booktabs}       
\usepackage{amsfonts}       
\usepackage{nicefrac}       
\usepackage{microtype}      
\usepackage{longtable}

\usepackage{amsmath,amssymb,amsthm}

\newcommand{\D}{\mathcal{D}}

\usepackage{algorithm}
\usepackage{algpseudocode}

\usepackage{graphicx}
\usepackage[caption=false]{subfig}

\usepackage{tabularx}

\usepackage{color}
\definecolor{dkgreen}{rgb}{0,0.6,0}
\definecolor{customgray}{rgb}{0.25,0.25,0.25}
\definecolor{customred}{rgb}{0.8,0.05,0.05}
\definecolor{customblue}{rgb}{0.05,0.05,0.8}

\usepackage{color}

\def\B{\mathbf{B}}
\def\C{\mathbf{C}}

\def\x{\mathbf{x}}
\def\m{\mathbf{m}}
\def\I{\mathbf{I}}
\def\v{\mathbf{v}}
\def\D{\mathbf{D}}
\def\O{\mathcal{O}}
\def\s{\mathbf{s}}
\def\tv{\mathbf{t}}
\def\m{\mathbf{m}}
\def\p{\mathbf{p}}
\def\t{\theta}
\def\F{\mathbf{F}}
\def\y{\mathbf{y}}
\def\z{\mathbf{z}}
\def\V{\mathbf{V}}
\def\0{\mathbf{0}}
\def\1{\mathbf{1}}
\def\H{\mathbf{H}}
\def\alphavd{\alpha_{\rm vd}}

%% file: main.bbl
\begin{thebibliography}{10}
\providecommand{\url}[1]{#1}
\csname url@samestyle\endcsname
\providecommand{\newblock}{\relax}
\providecommand{\bibinfo}[2]{#2}
\providecommand{\BIBentrySTDinterwordspacing}{\spaceskip=0pt\relax}
\providecommand{\BIBentryALTinterwordstretchfactor}{4}
\providecommand{\BIBentryALTinterwordspacing}{\spaceskip=\fontdimen2\font plus
\BIBentryALTinterwordstretchfactor\fontdimen3\font minus
  \fontdimen4\font\relax}
\providecommand{\BIBforeignlanguage}[2]{{%
\expandafter\ifx\csname l@#1\endcsname\relax
\typeout{** WARNING: IEEEtran.bst: No hyphenation pattern has been}%
\typeout{** loaded for the language `#1'. Using the pattern for}%
\typeout{** the default language instead.}%
\else
\language=\csname l@#1\endcsname
\fi
#2}}
\providecommand{\BIBdecl}{\relax}
\BIBdecl

\bibitem{feurer2019hyperparameter}
M.~Feurer and F.~Hutter, ``Hyperparameter optimization,'' in \emph{Automated
  machine learning}.\hskip 1em plus 0.5em minus 0.4em\relax Springer, Cham,
  2019, pp. 3--33.

\bibitem{bergstra2011algorithms}
J.~Bergstra, R.~Bardenet, Y.~Bengio, and B.~K{\'e}gl, ``{Algorithms for
  Hyper-Parameter Optimization},'' \emph{Advances in neural information
  processing systems}, vol.~24, 2011.

\bibitem{bergstra2012random}
J.~Bergstra and Y.~Bengio, ``{Random Search for Hyper-Parameter
  Optimization},'' \emph{Journal of machine learning research}, vol.~13, no.~2,
  2012.

\bibitem{turner2021bayesian}
R.~Turner, D.~Eriksson, M.~McCourt, J.~Kiili, E.~Laaksonen, Z.~Xu, and
  I.~Guyon, ``{Bayesian Optimization is Superior to Random Search for Machine
  Learning Hyperparameter Tuning: Analysis of the Black-Box Optimization
  Challenge 2020},'' \emph{arXiv preprint arXiv:2104.10201}, 2021.

\bibitem{frazier2018tutorial}
P.~I. Frazier, ``{A Tutorial on Bayesian Optimization},'' \emph{arXiv preprint
  arXiv:1807.02811}, 2018.

\bibitem{loshchilov2016cma}
I.~Loshchilov and F.~Hutter, ``{CMA-ES for Hyperparameter Optimization of Deep
  Neural Networks},'' \emph{arXiv preprint arXiv:1604.07269}, 2016.

\bibitem{nomura2021warm}
M.~Nomura, S.~Watanabe, Y.~Akimoto, Y.~Ozaki, and M.~Onishi, ``{Warm Starting
  CMA-ES for Hyperparameter Optimization},'' in \emph{Proceedings of the AAAI
  Conference on Artificial Intelligence}, vol.~35, no.~10, 2021, pp.
  9188--9196.

\bibitem{glasmachers2010exponential}
T.~Glasmachers, T.~Schaul, S.~Yi, D.~Wierstra, and J.~Schmidhuber,
  ``{Exponential Natural Evolution Strategies},'' in \emph{Proceedings of the
  12th annual conference on Genetic and evolutionary computation}.\hskip 1em
  plus 0.5em minus 0.4em\relax ACM, 2010, pp. 393--400.

\bibitem{wierstra2014natural}
D.~Wierstra, T.~Schaul, T.~Glasmachers, Y.~Sun, J.~Peters, and J.~Schmidhuber,
  ``{Natural Evolution Strategies},'' \emph{The Journal of Machine Learning
  Research}, vol.~15, no.~1, pp. 949--980, 2014.

\bibitem{mcgrath2016unscented}
C.~McGrath, M.~Karpenko, R.~Proulx, and I.~M. Ross, ``{An Unscented Natural
  Evolution Strategy for Solving Trajectory Optimization Problems},'' in
  \emph{AIAA/AAS Astrodynamics Specialist Conference}, 2016, p. 5512.

\bibitem{schaul2011high}
T.~Schaul, T.~Glasmachers, and J.~Schmidhuber, ``{High Dimensions and Heavy
  Tails for Natural Evolution Strategies},'' in \emph{Proceedings of the 13th
  annual conference on Genetic and evolutionary computation}.\hskip 1em plus
  0.5em minus 0.4em\relax ACM, 2011, pp. 845--852.

\bibitem{sun2009efficient}
Y.~Sun, D.~Wierstra, T.~Schaul, and J.~Schmidhuber, ``{Efficient Natural
  Evolution Strategies},'' in \emph{Proceedings of the 11th Annual conference
  on Genetic and evolutionary computation}.\hskip 1em plus 0.5em minus
  0.4em\relax ACM, 2009, pp. 539--546.

\bibitem{yi2009stochastic}
S.~Yi, D.~Wierstra, T.~Schaul, and J.~Schmidhuber, ``{Stochastic Search Using
  the Natural Gradient},'' in \emph{Proceedings of the 26th Annual
  International Conference on Machine Learning}.\hskip 1em plus 0.5em minus
  0.4em\relax ACM, 2009, pp. 1161--1168.

\bibitem{cuccu2012block}
G.~Cuccu and F.~Gomez, ``{Block Diagonal Natural Evolution Strategies},'' in
  \emph{International Conference on Parallel Problem Solving from
  Nature}.\hskip 1em plus 0.5em minus 0.4em\relax Springer, 2012, pp. 488--497.

\bibitem{bensadon2015black}
J.~Bensadon, ``{Black-Box Optimization Using Geodesics in Statistical
  Manifolds},'' \emph{Entropy}, vol.~17, no.~1, pp. 304--345, 2015.

\bibitem{li2020evolution}
Z.~Li, X.~Lin, Q.~Zhang, and H.~Liu, ``Evolution strategies for continuous
  optimization: A survey of the state-of-the-art,'' \emph{Swarm and
  Evolutionary Computation}, vol.~56, p. 100694, 2020.

\bibitem{glasmachers2010natural}
T.~Glasmachers, T.~Schaul, and J.~Schmidhuber, ``{A Natural Evolution Strategy
  for Multi-Objective Optimization},'' in \emph{International Conference on
  Parallel Problem Solving from Nature}.\hskip 1em plus 0.5em minus 0.4em\relax
  Springer, 2010, pp. 627--636.

\bibitem{schaul2012natural}
T.~Schaul, ``Natural evolution strategies converge on sphere functions,'' in
  \emph{Proceedings of the 14th annual conference on Genetic and evolutionary
  computation}, 2012, pp. 329--336.

\bibitem{nomura2022towards}
M.~Nomura and I.~Ono, ``Towards a principled learning rate adaptation for
  natural evolution strategies,'' in \emph{International Conference on the
  Applications of Evolutionary Computation (Part of EvoStar)}.\hskip 1em plus
  0.5em minus 0.4em\relax Springer, 2022, pp. 721--737.

\bibitem{beyer2014convergence}
H.-G. Beyer, ``{Convergence Analysis of Evolutionary Algorithms That Are Based
  on the Paradigm of Information Geometry},'' \emph{Evolutionary Computation},
  vol.~22, no.~4, pp. 679--709, 2014.

\bibitem{muller2021non}
N.~M{\"u}ller and T.~Glasmachers, ``{Non-Local Optimization: Imposing Structure
  on Optimization Problems by Relaxation},'' in \emph{Proceedings of the 16th
  ACM/SIGEVO Conference on Foundations of Genetic Algorithms}, 2021, pp. 1--10.

\bibitem{amari1998natural}
S.-I. Amari, ``{Natural Gradient Works Efficiently in Learning},'' \emph{Neural
  computation}, vol.~10, no.~2, pp. 251--276, 1998.

\bibitem{amari1998why}
S.-I. Amari and S.~C. Douglas, ``{Why natural gradient?}'' in \emph{Proceedings
  of the 1998 IEEE International Conference on Acoustics, Speech and Signal
  Processing, ICASSP'98 (Cat. No. 98CH36181)}, vol.~2.\hskip 1em plus 0.5em
  minus 0.4em\relax IEEE, 1998, pp. 1213--1216.

\bibitem{pascanu2013revisiting}
R.~Pascanu and Y.~Bengio, ``Revisiting natural gradient for deep networks,''
  \emph{arXiv preprint arXiv:1301.3584}, 2013.

\bibitem{martens2020new}
J.~Martens, ``New insights and perspectives on the natural gradient method,''
  \emph{Journal of Machine Learning Research}, vol.~21, no. 146, pp. 1--76,
  2020.

\bibitem{kullback1951information}
S.~Kullback and R.~A. Leibler, ``{On Information and Sufficiency},'' \emph{The
  annals of mathematical statistics}, vol.~22, no.~1, pp. 79--86, 1951.

\bibitem{malago2015information}
L.~Malag{\`o} and G.~Pistone, ``{Information Geometry of the Gaussian
  Distribution in View of Stochastic Optimization},'' in \emph{Proceedings of
  the 2015 ACM Conference on Foundations of Genetic Algorithms XIII}, 2015, pp.
  150--162.

\bibitem{ollivier2017information}
Y.~Ollivier, L.~Arnold, A.~Auger, and N.~Hansen, ``{Information-Geometric
  Optimization Algorithms: A Unifying Picture via Invariance Principles},''
  \emph{The Journal of Machine Learning Research}, vol.~18, no.~1, pp.
  564--628, 2017.

\bibitem{nomura2021natural}
M.~Nomura and I.~Ono, ``Natural evolution strategy for unconstrained and
  implicitly constrained problems with ridge structure,'' in \emph{2021 IEEE
  Symposium Series on Computational Intelligence (SSCI)}.\hskip 1em plus 0.5em
  minus 0.4em\relax IEEE, 2021, pp. 1--7.

\bibitem{nomura2021distance}
M.~Nomura, N.~Sakai, N.~Fukushima, and I.~Ono, ``Distance-weighted exponential
  natural evolution strategy for implicitly constrained black-box function
  optimization,'' in \emph{2021 IEEE Congress on Evolutionary Computation
  (CEC)}.\hskip 1em plus 0.5em minus 0.4em\relax IEEE, 2021, pp. 1099--1106.

\bibitem{hansen2003reducing}
N.~Hansen, S.~D. M{\"u}ller, and P.~Koumoutsakos, ``{Reducing the Time
  Complexity of the Derandomized Evolution Strategy with Covariance Matrix
  Adaptation (CMA-ES)},'' \emph{Evolutionary computation}, vol.~11, no.~1, pp.
  1--18, 2003.

\bibitem{hansen2001completely}
N.~Hansen and A.~Ostermeier, ``{Completely Derandomized Self-Adaptation in
  Evolution Strategies},'' \emph{Evolutionary computation}, vol.~9, no.~2, pp.
  159--195, 2001.

\bibitem{hansen2006cma}
N.~Hansen, ``{The CMA Evolution Strategy: A Comparing Review},'' \emph{Towards
  a new evolutionary computation}, pp. 75--102, 2006.

\bibitem{hansen2016cma}
------, ``{The CMA Evolution Strategy: A Tutorial},'' \emph{arXiv preprint
  arXiv:1604.00772}, 2016.

\bibitem{akimoto2014comparison}
Y.~Akimoto, A.~Auger, and N.~Hansen, ``{Comparison-Based Natural Gradient
  Optimization in High Dimension},'' in \emph{Proceedings of the 2014 Annual
  Conference on Genetic and Evolutionary Computation}.\hskip 1em plus 0.5em
  minus 0.4em\relax ACM, 2014, pp. 373--380.

\bibitem{fukushima2011proposal}
N.~Fukushima, Y.~Nagata, S.~Kobayashi, and I.~Ono, ``{Proposal of
  distance-weighted exponential natural evolution strategies},'' in \emph{2011
  IEEE Congress of Evolutionary Computation (CEC)}.\hskip 1em plus 0.5em minus
  0.4em\relax IEEE, 2011, pp. 164--171.

\bibitem{hansen1996adapting}
N.~Hansen and A.~Ostermeier, ``{Adapting Arbitrary Normal Mutation
  Distributions in Evolution Strategies: The Covariance Matrix Adaptation},''
  in \emph{Proceedings of IEEE international conference on evolutionary
  computation}.\hskip 1em plus 0.5em minus 0.4em\relax IEEE, 1996, pp.
  312--317.

\bibitem{ros2008simple}
R.~Ros and N.~Hansen, ``{A Simple Modification in CMA-ES Achieving Linear Time
  and Space Complexity},'' in \emph{International Conference on Parallel
  Problem Solving from Nature}.\hskip 1em plus 0.5em minus 0.4em\relax
  Springer, 2008, pp. 296--305.

\bibitem{sun2013linear}
Y.~Sun, T.~Schaul, F.~Gomez, and J.~Schmidhuber, ``{A Linear Time Natural
  Evolution Strategy for Non-Separable Functions},'' in \emph{Proceedings of
  the 15th annual conference companion on Genetic and evolutionary
  computation}, 2013, pp. 61--62.

\bibitem{auger2005restart}
A.~Auger and N.~Hansen, ``{A Restart CMA Evolution Strategy With Increasing
  Population Size},'' in \emph{2005 IEEE congress on evolutionary computation},
  vol.~2.\hskip 1em plus 0.5em minus 0.4em\relax IEEE, 2005, pp. 1769--1776.

\bibitem{akimoto2016online}
Y.~Akimoto and N.~Hansen, ``{Online Model Selection for Restricted Covariance
  Matrix Adaptation},'' in \emph{International Conference on Parallel Problem
  Solving from Nature}.\hskip 1em plus 0.5em minus 0.4em\relax Springer, 2016,
  pp. 3--13.

\bibitem{sun2019survey}
S.~Sun, Z.~Cao, H.~Zhu, and J.~Zhao, ``{A Survey of Optimization Methods from a
  Machine Learning Perspective},'' \emph{IEEE transactions on cybernetics},
  vol.~50, no.~8, pp. 3668--3681, 2019.

\bibitem{sra2012optimization}
S.~Sra, S.~Nowozin, and S.~J. Wright, \emph{{Optimization for Machine
  Learning}}.\hskip 1em plus 0.5em minus 0.4em\relax Mit Press, 2012.

\bibitem{li2018visualizing}
H.~Li, Z.~Xu, G.~Taylor, C.~Studer, and T.~Goldstein, ``{Visualizing the Loss
  Landscape of Neural Nets},'' \emph{Advances in Neural Information Processing
  Systems}, vol.~31, 2018.

\bibitem{bottou2012stochastic}
L.~Bottou, ``{Stochastic Gradient Descent Tricks},'' in \emph{Neural networks:
  Tricks of the trade}.\hskip 1em plus 0.5em minus 0.4em\relax Springer, 2012,
  pp. 421--436.

\bibitem{ruder2016overview}
S.~Ruder, ``{An Overview of Gradient Descent Optimization Algorithms},''
  \emph{arXiv preprint arXiv:1609.04747}, 2016.

\end{thebibliography}
